\documentclass[letterpaper, 10 pt, conference]{ieeeconf}

\IEEEoverridecommandlockouts
\usepackage{graphicx}
\usepackage{amsmath}
\usepackage{amssymb}
\usepackage{booktabs}
\usepackage{multirow}
\usepackage{url}
\usepackage{cite}
\usepackage[table]{xcolor}
\usepackage{array}

\usepackage{pifont}
\usepackage[pagebackref,breaklinks,colorlinks,allcolors=DeepPink]{hyperref}

\definecolor{ICRABlue}{rgb}{0.21,0.49,0.74}
\definecolor{DeepPink}{rgb}{1,0.078,0.576}
\definecolor{Gray}{rgb}{0.4, 0.4, 0.4}

\title{\LARGE \bf
\raisebox{-0.18\height}{\includegraphics[height=1.25em]{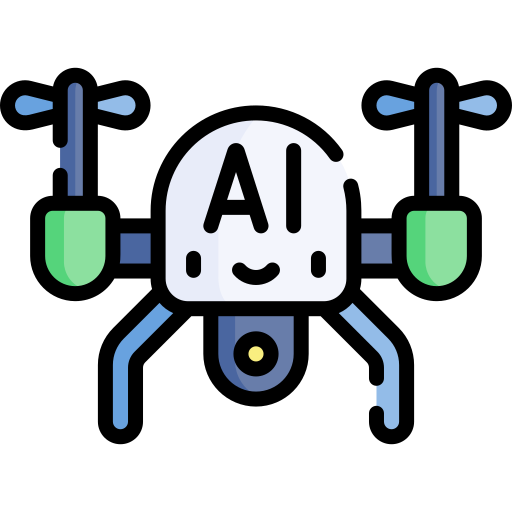}}\hspace{0.25em}%
UAV-DualCog: A Dual-Cognition Benchmark for Grounded UAV Agent-World Reasoning with MLLMs
}

\author{Like Liu$^{1}$, Zhengzheng Xu$^{2}$, Hongzhe Li$^{2}$, Lin Wang$^{3}$, and Dian Shao$^{2,\dagger}$%
\thanks{$^{1}$Like Liu is with the School of Software, Northwestern Polytechnical University, Xi'an, Shaanxi, China (like.liu@mail.nwpu.edu.cn).}%
\thanks{$^{2}$Zhengzheng Xu, Hongzhe Li, and Dian Shao are with the Unmanned System Research Institute and the School of Artificial Intelligence, Northwestern Polytechnical University, Xi'an, Shaanxi, China.}%
\thanks{$^{3}$Lin Wang is with the Embodied Perception and Interaction Lab (EmPACT Lab), School of Electrical and Electronic Engineering, Nanyang Technological University, Singapore.}%
\thanks{$^{\dagger}$Corresponding author: Dian Shao (shaodian@nwpu.edu.cn).}%
}

\IEEEaftertitletext{\begingroup
    \centering
    \setlength{\abovecaptionskip}{2pt}
    \includegraphics[width=\textwidth]{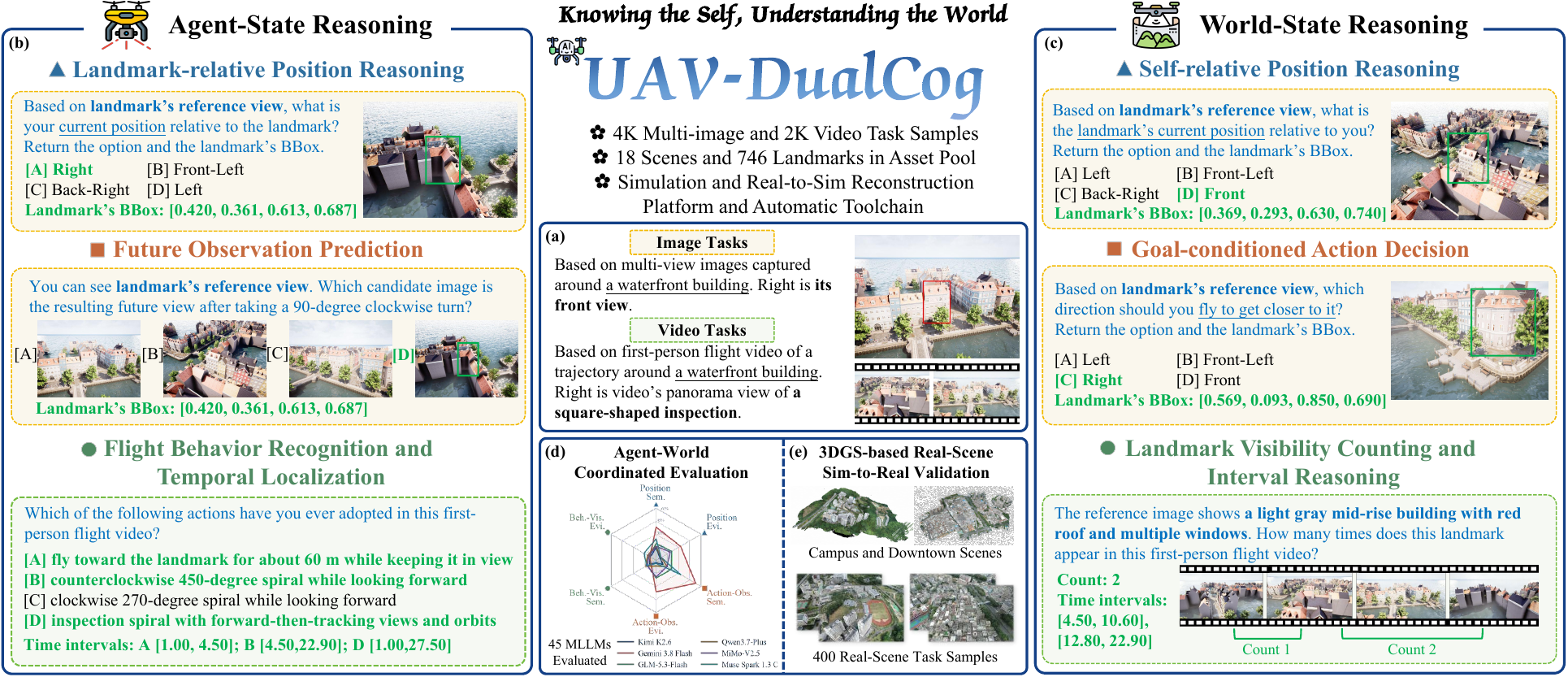}
    \vspace{-12pt}
    \refstepcounter{figure}\label{fig:benchmark_overview}
    \makeatletter
    \@makecaption{Fig.~\thefigure}{\textbf{UAV-DualCog} benchmarks grounded \textbf{\emph{UAV dual cognition}} by testing whether a UAV agent can coordinate its own state with the surrounding world and localize the evidence across space and time. (a)~Matched multi-view images and egocentric videos define Agent--Landmark contexts. (b)~\textbf{\emph{Agent-State Reasoning}} covers the UAV's relative position, motion, behavior, and action--observation relations. (c)~\textbf{\emph{World-State Reasoning}} covers landmark position, visibility, and spatial relations; BBoxes and intervals ground semantic answers. (d)~The radar compares MLLMs on paired semantic and evidence co-success across coordinated abilities: \textcolor[HTML]{2F6F9F}{\ding{115}} relative-position, \textcolor[HTML]{C66A3D}{\ding{110}} action--observation, and \textcolor[HTML]{4F8A62}{\ding{108}} behavior--visibility reasoning. (e)~Sim-to-real validation supports findings beyond simulation environments.}
    \vspace{2pt}
    \makeatother
    \par
\endgroup
}

\begin{document}

\maketitle
\thispagestyle{empty}
\pagestyle{empty}

%%%%%%%%%%%%%%%%%%%%%%%%%%%%%%%%%%%%%%%%%%%%%%%%%%%%%%%%%%%%%%%%%%%%%%%%%%%%%%%%
\begin{abstract}
Multimodal large language models (MLLMs) perform strongly across vision-language tasks, yet remain insufficiently explored in UAV scenarios. Recent UAV-oriented benchmarks have begun to evaluate MLLMs in aerial scenarios, but typically focus on scene understanding, event recognition, or navigation completion, rather than jointly assessing the dual-cognition capability required for UAV agents: reasoning about the UAV's own position, behavior, and observation state alongside external landmark states under continuously changing viewpoints, supported by explicit spatio-temporal grounding.
To address this gap, we present \textbf{UAV-DualCog}, the first benchmark to formulate UAV dual cognition as the coordinated evaluation of \emph{Agent-State Reasoning} and \emph{World-State Reasoning} across multi-view images and egocentric videos, with explicit spatial or temporal grounding. The State-aware UAV Simulation Platform (SUSP) and the DualCog-aware Capture System (DCCS) construct 4,096 image and 2,048 video samples from 12 scenes, pairing complementary state questions within shared Agent--Landmark contexts.
Extensive evaluations on 45 models show that current MLLMs remain far from reliable in UAV dual cognition: Agent-State Reasoning is generally weaker than World-State Reasoning, semantic answers do not reliably coincide with spatial or temporal grounding, and coordinated abilities remain limited. Thinking-mode comparisons, open-ended analyses, and sim-to-real validation show that these limitations persist across inference settings, formats, and environments.
We further construct UAV-DualCog-Train and improve UAV dual-cognition performance through a lightweight optimization probe. Supplementary website, code, and dataset: \url{https://smartdianlab.github.io/UAV-DualCog/}.

\end{abstract}

%%%%%%%%%%%%%%%%%%%%%%%%%%%%%%%%%%%%%%%%%%%%%%%%%%%%%%%%%%%%%%%%%%%%%%%%%%%%%%%%
\section{Introduction}

\begin{table*}[!t]
\centering
\caption{Comparison with representative aerial and embodied reasoning benchmarks. Checkmarks denote explicit protocol targets. Paired Eval. uses sample-aligned Agent-State and World-State questions for the same landmark, view, or trajectory. Action--Obs. Reasoning covers action-induced observation prediction and observation-induced action decisions.}
\label{tab:related_benchmarks}

\scriptsize
\setlength{\tabcolsep}{1.7pt}
\renewcommand{\arraystretch}{1.08}

\providecommand{\cmark}{\textcolor{green!60!black}{\ding{51}}}
\providecommand{\xmark}{\textcolor{red!70!black}{\ding{55}}}

\begin{tabular}{@{}
>{\raggedright\arraybackslash}m{0.18\textwidth}|
>{\centering\arraybackslash}m{0.06\textwidth}
>{\centering\arraybackslash}m{0.08\textwidth}
>{\centering\arraybackslash}m{0.075\textwidth}|
>{\centering\arraybackslash}m{0.075\textwidth}
>{\centering\arraybackslash}m{0.065\textwidth}
>{\centering\arraybackslash}m{0.075\textwidth}|
>{\centering\arraybackslash}m{0.05\textwidth}|
>{\centering\arraybackslash}m{0.06\textwidth}
>{\centering\arraybackslash}m{0.075\textwidth}|
>{\centering\arraybackslash}m{0.05\textwidth}
>{\centering\arraybackslash}m{0.05\textwidth}@{}}
\toprule
\multicolumn{1}{c|}{\multirow{2}{*}{\textbf{Benchmark}}}
& \multicolumn{3}{c|}{\textbf{Agent-State}}
& \multicolumn{3}{c|}{\textbf{World-State}}
& \textbf{Paired}
& \multicolumn{2}{c|}{\textbf{Explicit Evidence}}
& \multicolumn{2}{c}{\textbf{Modality}} \\
\cmidrule(l{2pt}r{2pt}){2-4}
\cmidrule(l{2pt}r{2pt}){5-7}
\cmidrule(l{2pt}r{2pt}){9-10}
\cmidrule(l{2pt}r{2pt}){11-12}
& \shortstack{\textbf{UAV}\\\textbf{Position}}
& \shortstack{\textbf{Motion /}\\\textbf{Behavior}}
& \shortstack{\textbf{Action--Obs.}\\\textbf{Reasoning}}
& \shortstack{\textbf{Landmark}\\\textbf{Attribute}}
& \shortstack{\textbf{Landmark}\\\textbf{Visibility}}
& \shortstack{\textbf{Spatial}\\\textbf{Relation}}
& \textbf{Eval}
& \textbf{BBox}
& \shortstack{\textbf{Temporal}\\\textbf{Interval}}
& \textbf{Image}
& \textbf{Video} \\
\midrule

\rowcolor{gray!15}
\multicolumn{12}{l}{\rule[-0.5ex]{0pt}{2.2ex}\textbf{\textit{Vision-Language Navigation}}} \\
AerialVLN~\cite{liu_2023_AerialVLN}
& \xmark & \xmark & \xmark & \xmark & \xmark & \xmark & \xmark & \xmark & \xmark & \cmark & \xmark \\
OpenFly~\cite{OpenFly}
& \xmark & \xmark & \xmark & \xmark & \xmark & \xmark & \xmark & \xmark & \xmark & \cmark & \xmark \\

\rowcolor{gray!15}
\multicolumn{12}{l}{\rule[-0.5ex]{0pt}{2.2ex}\textbf{\textit{Vision-Language-Action and Aerial Object Search}}} \\
HUGE-Bench~\cite{guo2026huge}
& \xmark & \xmark & \xmark & \xmark & \xmark & \xmark & \xmark & \xmark & \xmark & \cmark & \xmark \\
OpenUAV~\cite{openuav}
& \xmark & \xmark & \xmark & \cmark & \cmark & \xmark & \xmark & \xmark & \xmark & \cmark & \xmark \\
CityAVOS~\cite{ji2026towards}
& \xmark & \xmark & \xmark & \cmark & \cmark & \xmark & \xmark & \xmark & \xmark & \cmark & \xmark \\

\rowcolor{gray!15}
\multicolumn{12}{l}{\rule[-0.5ex]{0pt}{2.2ex}\textbf{\textit{Embodied Question Answering}}} \\
CityEQA~\cite{zhao2025cityeqa}
& \xmark & \xmark & \xmark & \cmark & \cmark & \xmark & \xmark & \xmark & \xmark & \cmark & \xmark \\
Open3D-VQA~\cite{zhang2025open3d}
& \xmark & \xmark & \xmark & \cmark & \cmark & \xmark & \xmark & \xmark & \xmark & \cmark & \xmark \\
UrbanVideo-Bench~\cite{zhao2025urbanvideobench}
& \cmark & \cmark & \cmark & \cmark & \cmark & \cmark & \xmark & \xmark & \xmark & \xmark & \cmark \\
CityCube~\cite{xu2026citycube}
& \cmark & \cmark & \cmark & \cmark & \cmark & \cmark & \xmark & \xmark & \xmark & \cmark & \xmark \\
SIS-Bench~\cite{zou2026selfinspace}
& \xmark & \cmark & \cmark & \cmark & \cmark & \xmark & \xmark & \xmark & \xmark & \xmark & \cmark \\
\midrule
\rowcolor{cyan!10}
{\scriptsize\bfseries UAV-DualCog (Ours)}
& \cmark & \cmark & \cmark & \cmark & \cmark & \cmark
& \cmark & \cmark & \cmark & \cmark & \cmark \\
\bottomrule
\end{tabular}
\end{table*}

Recent advances in multimodal large language models (MLLMs) have greatly improved machine performance on perception and reasoning tasks that combine visual and linguistic information~\cite{yang2025thinking,li2025perception,liang2026hispatial}. As these models are increasingly considered for embodied agents~\cite{zeng2023large,lin2024embodied,shao2026foundationuav,shao2026vlnsurvey}, an important question is whether they can support reliable reasoning in UAV scenarios, where perception is shaped by continuous motion and rapidly changing viewpoints in open 3D space~\cite{liu_2023_AerialVLN,jiang2025longfly,wu2025aeroduo,lee2025citynav}. Reliable agents must determine not only what exists in the scene, but also where they are, how they are moving, and how their actions relate to future observations. These capabilities directly support navigation subgoal completion, relative-position arrival, viewpoint adjustment, and sustained landmark tracking and reacquisition.

A key challenge in UAV embodied intelligence is that the agent must reason not only about the external environment, but also about its own state during motion. We call this coordinated reasoning \emph{dual cognition}. \emph{Agent-State Reasoning} covers the UAV's relative position, motion, behavior, and action--observation relations; \emph{World-State Reasoning} covers landmark position, visibility, and spatial relations. Pairing these complementary views within the same Agent--Landmark world context assesses capabilities needed for coordinate-frame reasoning, action--observation association, and behavior--visibility localization. Existing aerial benchmarks cover scene understanding~\cite{zhao2025urbanvideobench,zhang2025open3d}, event recognition~\cite{li2021uav}, spatial reasoning~\cite{miyanishi_2023_NeurIPS,zhang2025open3d}, embodied question answering~\cite{guo2026huge,zhao2025cityeqa}, and navigation~\cite{liu_2023_AerialVLN,openuav,lee2025citynav,OpenFly,shao2026finecognav}; few jointly test both states across multi-view images and egocentric videos while requiring explicit spatial and temporal grounding.

To address this gap, we introduce \textbf{UAV-DualCog}, shown in Fig.~\ref{fig:benchmark_overview}. To our knowledge, UAV-DualCog is the first benchmark to formulate UAV dual cognition as the coordinated evaluation of UAV embodied Agent-State and World-State Reasoning across multi-view images and egocentric videos, with explicit spatio-temporal grounding.
The released benchmark spans 6 grounded tasks, containing 4,096 image and 2,048 video samples from 12 scenes and 512 landmarks. To ensure scalability and geometric consistency, we develop an automated construction pipeline: the State-aware UAV Simulation Platform (SUSP) consolidates geometric state-aware scene scanning into a reusable multi-view landmark asset pool, while the DualCog-aware Capture System (DCCS) automates collision-aware trajectory planning, deriving paired questions and bounding-box or temporal-interval annotations from synchronized state records and geometric projection.

Based on this benchmark, we evaluate 45 proprietary, open-weight, and spatially specialized MLLMs. Results reveal persistent gaps between semantic prediction and evidence grounding, weaker Agent-State Reasoning, and unstable temporal integration. Human evaluation, thinking/frontier-model comparisons, failure diagnosis, and open-ended testing further characterize these limitations. A pilot with 30 landmarks and 400 image questions from 3 real-scene 3DGS reconstructions reproduces the main capability separation. We also construct UAV-DualCog-Train with 234 landmarks and 11,232 samples from 6 disjoint scenes; a lightweight two-stage optimization probe shows that the benchmark design can provide structured supervision.

Our contributions are summarized as follows:
\begin{itemize}
    \item We introduce \textbf{UAV-DualCog}, a dual-cognition benchmark that jointly evaluates Agent-State and World-State Reasoning across image and video observations with explicit spatial and temporal grounding.
    \item We develop the State-aware UAV Simulation Platform (SUSP) and the DualCog-aware Capture System (DCCS), which scale six grounded reasoning tasks from verifiable pose, landmark, visibility, and trajectory records.
    \item We benchmark diverse MLLMs, analyze semantic--evidence failures and coordinated abilities, and verify with UAV-DualCog-Train that the data can provide useful structured supervision.
\end{itemize}

\section{Related Work}

\begin{figure*}[t]
    \centering
    \includegraphics[width=\linewidth]{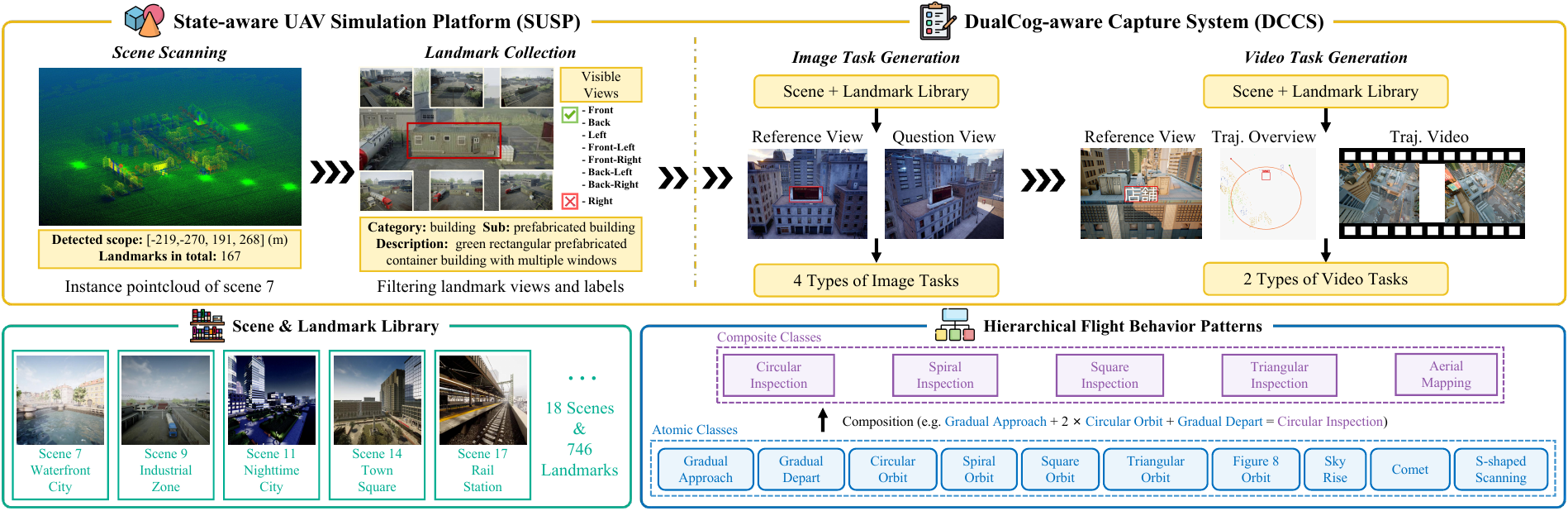}
    \caption{Two-layer construction system. The State-aware UAV Simulation Platform (SUSP) consolidates scene scanning and landmark asset construction; the DualCog-aware Capture System (DCCS) contains image and video task modules for grounded sample generation.}
    \label{fig:benchmark_construction_process}
\end{figure*}

\subsection{MLLM Benchmarks and Evaluation}
Recent MLLMs have shown strong performance across a wide range of vision-language benchmarks. Existing evaluations mainly assess visual question answering, grounding, spatial understanding, and multi-step reasoning~\cite{fu2023mme,liu2024mmbench,li2024seed,yu2023mm,yu2024mm,yue2024mmmu,ge2025mllm}. However, these benchmarks are largely built on static images or short videos in passive observation settings, where inputs are not conditioned on an embodied agent's motion or viewpoint. Consequently, strong benchmark accuracy does not establish whether a model can track its own state, interpret an action-conditioned observation, or localize relevant objects and events during movement.

\subsection{UAV-Oriented Benchmarks}
Existing UAV benchmarks focus primarily on aerial perception and tracking~\cite{barekatain2017okutama,li2021uav,du2018unmanned}, spatial reasoning~\cite{zhang2025open3d}, embodied question answering~\cite{zhao2025cityeqa,guo2026huge}, or navigation~\cite{liu_2023_AerialVLN,zhang2025logisticsvln,OpenFly,shao2026finecognav}. These efforts remain largely environment-centric, evaluating scene understanding or goal completion without explicitly modeling the agent's state. Concurrent SIS-Bench~\cite{zou2026selfinspace} introduces self-awareness into UAV evaluation and organizes 13 real-video tasks along self-awareness, spatial cognition, and perception--memory--reasoning levels. Its independently evaluated multiple-choice tasks diagnose a broad cognitive hierarchy, but do not report paired Agent-State/World-State success for a matched Agent--Landmark context or require BBox and interval outputs. UAV-DualCog instead formulates dual cognition across multi-view images and egocentric videos: matched image strata and shared video trajectories pair complementary state questions, while spatial and temporal outputs separately evaluate landmark and event localization. State-aware simulation provides complete 6-DoF pose, geometry, and visibility metadata for controlled construction, while a separate real-scene 3DGS pilot tests protocol portability beyond simulation environments. Table~\ref{tab:related_benchmarks} summarizes these protocol-level differences.

\section{The UAV-DualCog Benchmark}

\subsection{Dual-Cognition Formulation}
UAV visual understanding differs from conventional image or video understanding because the observation is generated by an \emph{active} aerial agent: its 3D pose, motion, behavior, and camera viewpoint directly affect visibility and scene interpretation. UAV-DualCog therefore treats these factors as core reasoning targets rather than background variables. At time $t$, we represent \emph{dual cognition} as the coupled agent--world state
\begin{equation}
\begin{aligned}
    \mathcal{D}_t &= (c_t^a,c_t^w), \\
    (\hat{y}_t^q,\hat{e}_t^q) &= f_\theta(o_{1:t},g,q),\quad q\in\{a,w\},
\end{aligned}
\end{equation}
where $c_t^a$ and $c_t^w$ are the Agent-State and World-State components; $o_{1:t}$, $g$, and $q$ denote the observations, referenced landmark or goal, and queried component; and $(\hat{y}_t^q,\hat{e}_t^q)$ are the predicted semantic state and grounding. Agent-State Reasoning covers the UAV's relative position, motion, behavior, and action--observation relations, whereas World-State Reasoning covers landmark position, visibility, and spatial relations. Paired records assess complementary questions within the same Agent--Landmark world context, sharing a scene, landmark, and trajectory.
This formulation supports paired assessment of complementary agent--world states.

\subsection{Grounded Dual-Cognition Tasks}
UAV-DualCog contains six tasks across two media. Four multi-view image tasks predict an answer option and a normalized landmark bounding box. \emph{Landmark-relative Position Reasoning} locates the UAV relative to a landmark, while \emph{Action-conditioned Future Observation Prediction} selects the view produced by an action; both target Agent-State Reasoning. \emph{Self-relative Position Reasoning} locates a landmark relative to the UAV, and \emph{Goal-conditioned Action Decision} selects an action for approaching it; both target World-State Reasoning. The paired relative-position tasks assess reciprocal interpretation across landmark-centric and UAV-centric frames, while the future-observation and action-decision pair assesses the bidirectional action--observation association.

Two egocentric video tasks require semantic predictions and temporal intervals. \emph{Flight Behavior Recognition and Temporal Localization} evaluates Agent-State Reasoning at composite and atomic behavior levels. \emph{Landmark Visibility Counting and Interval Reasoning} evaluates World-State Reasoning by tracking target landmark appearances.
Semantic and localization outputs are evaluated independently using the metrics in Sec.~IV-B. Task definitions are summarized on the Benchmark page; full prompts and output contracts are on the Evaluation page of the supplementary website.

\subsection{State-aware and Scalable Construction System}
The construction system is organized into two layers (Fig.~\ref{fig:benchmark_construction_process}). The \textbf{State-aware UAV Simulation Platform (SUSP)} consolidates scene scanning and landmark asset construction. It extends AirSim with unified APIs for 6-DoF pose, camera intrinsics/extrinsics, semantic point clouds, landmark geometry, and collisions. Coverage-aware pose sampling synchronizes LiDAR, RGB, segmentation, and telemetry. Multi-view spatial association and 3D instance aggregation then cluster candidate landmarks for geometric/semantic annotation and human verification, producing reusable multi-view assets.

The \textbf{DualCog-aware Capture System (DCCS)} generates grounded multimodal dual-cognition tasks through two modules.
\ding{118} The \emph{Video Module} instantiates hierarchical flight behaviors (coupling high-level navigation patterns with atomic flight maneuvers), validates kinematically feasible and collision-free trajectories, records first-person video, and extracts synchronized behavior phases and landmark visibility intervals.
\ding{118} The \emph{Image Module} samples controlled UAV--landmark viewpoints across distances and viewing angles, validates camera-frustum projections, and derives canonical answers and 2D boxes. Views reused from the SUSP asset pool retain their reviewed boxes, which enclose the projected visible landmark points. Both modules apply schema and geometric sanity checks. Further API, synchronization, trajectory, and review workflow details are on the Construction page.

\subsection{Benchmark Statistics}
\begin{figure}[t]
    \centering
    \includegraphics[width=1\linewidth]{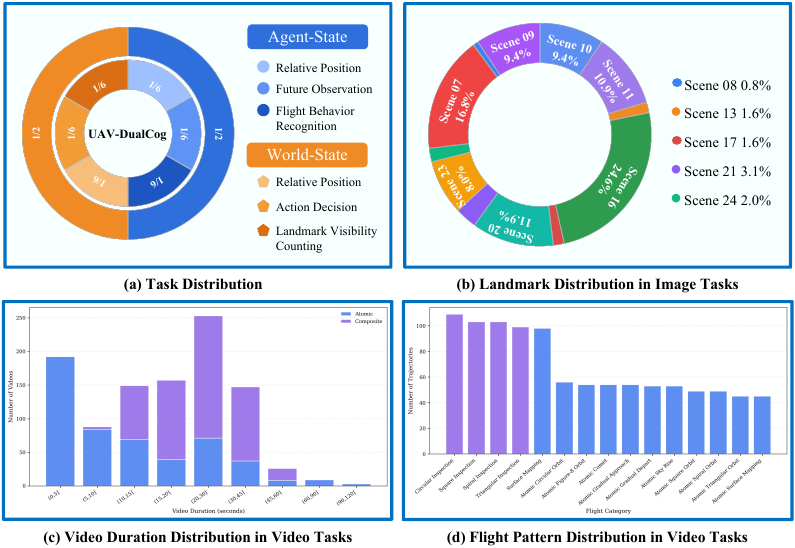}
    \caption{Statistics of the UAV-DualCog benchmark, showing the balanced task distribution, the landmark distribution across released scenes, the video duration distribution, and the flight pattern distribution in video tasks.}
    \label{fig:benchmark_stat}
\end{figure}

The SUSP asset pool spans 18 scenarios ($5.04\,\text{km}^2$) and retains 746 landmarks from 2,626 candidates. The released benchmark uses 12 scenarios and 512 landmarks to construct 4,096 image and 2,048 video samples balanced across six tasks. The video subset is built from 1,024 trajectories totaling 19,633.30 seconds. Fig.~\ref{fig:benchmark_stat} summarizes the statistics.

\section{Experiments}
% Queue the two full-width result tables early so they can occupy the tops of
% the following pages in two-column mode.
\definecolor{bestblue}{HTML}{BBEBFF}
\definecolor{secondblue}{HTML}{E6F5FF}

\begin{table*}[t]
\centering
\caption{Image-task results (\%). Dual columns aggregate matched scene--Agent--Landmark--difficulty pairs. \textsuperscript{T}: thinking-only endpoint at low effort (minimum effort available for Muse Spark). \colorbox{bestblue}{Dark} and \colorbox{secondblue}{light} blue mark the best and second-best model within each group.}
\label{tab:results_image}
\setlength{\tabcolsep}{3pt} 
\scriptsize
\setlength{\tabcolsep}{1.8pt}
\begin{tabular}{l|ccc|ccc|ccc|ccc|cc}
\toprule
\multirow{3}{*}{\textbf{Model}} & \multicolumn{6}{c|}{\textbf{Agent-State Reasoning}} & \multicolumn{6}{c|}{\textbf{World-State Reasoning}} & \multicolumn{2}{c}{\textbf{Coordination}} \\
\cmidrule(lr){2-7} \cmidrule(lr){8-13} \cmidrule(lr){14-15}
& \multicolumn{3}{c|}{\textbf{Rel. Position}} & \multicolumn{3}{c|}{\textbf{Future Obs.}} & \multicolumn{3}{c|}{\textbf{Rel. Position}} & \multicolumn{3}{c|}{\textbf{Action Dec.}} & \multirow{2}{*}{Dual Sem.} & \multirow{2}{*}{Dual Evi.} \\
& Acc & Acc@0.5 & mIoU  & Acc & Acc@0.5 & mIoU  & Acc & Acc@0.5 & mIoU  & Acc & Acc@0.5 & mIoU & & \\
\midrule
\rowcolor{gray!15} \multicolumn{15}{l}{\textit{\textbf{Proprietary Models}}} \\
Claude Opus 4.8\textsuperscript{T}\cite{anthropic2026claudeopus48} & 39.4\% & 11.4\% & 23.5\% & 35.9\% & 14.3\% & 25.4\% & 54.1\% & 10.4\% & 23.7\% & 51.9\% & 8.0\% & 21.7\% & 19.5\% & 2.5\% \\
Claude Opus 4.7\textsuperscript{T}\cite{anthropic2026claudeopus47} & 39.3\% & 8.2\% & 20.1\% & 37.2\% & 7.0\% & 18.1\% & 49.7\% & 7.1\% & 18.6\% & 39.1\% & 7.9\% & 18.0\% & 16.7\% & 1.1\% \\
Claude Opus 4.6\textsuperscript{T}\cite{anthropic2026claudeopus46} & 38.1\% & 2.1\% & 15.6\% & 32.4\% & 2.8\% & 18.7\% & 62.0\% & 8.6\% & 19.7\% & 58.9\% & 9.2\% & 20.1\% & 20.5\% & 0.0\% \\
Claude Sonnet 4.6\cite{anthropic2026claude46} & 37.7\% & 4.9\% & 18.2\% & 23.6\% & 3.6\% & 15.1\% & 48.5\% & 9.6\% & 20.2\% & 61.0\% & 8.6\% & 19.9\% & 15.9\% & 0.1\% \\
Gemini 3.8 Flash\textsuperscript{T}\cite{deepmind2026gemini38flash} & \cellcolor{bestblue}{60.1\%} & \cellcolor{bestblue}{66.7\%} & \cellcolor{bestblue}{57.6\%} & \cellcolor{bestblue}{74.9\%} & \cellcolor{bestblue}{73.8\%} & \cellcolor{bestblue}{62.5\%} & 61.9\% & \cellcolor{bestblue}{40.8\%} & \cellcolor{bestblue}{39.0\%} & \cellcolor{bestblue}{69.5\%} & \cellcolor{bestblue}{42.3\%} & \cellcolor{bestblue}{39.3\%} & \cellcolor{bestblue}{44.4\%} & \cellcolor{bestblue}{34.3\%} \\
Gemini 3.1 Pro Preview\textsuperscript{T}\cite{deepmind2026gemini31pro} & 40.9\% & \cellcolor{secondblue}{58.8\%} & \cellcolor{secondblue}{50.5\%} & \cellcolor{secondblue}{58.8\%} & \cellcolor{secondblue}{59.9\%} & \cellcolor{secondblue}{52.3\%} & 63.7\% & \cellcolor{secondblue}{37.5\%} & \cellcolor{secondblue}{36.8\%} & 47.2\% & \cellcolor{secondblue}{37.1\%} & \cellcolor{secondblue}{36.4\%} & 27.1\% & \cellcolor{secondblue}{30.1\%} \\
Gemini 3.1 Flash Lite\cite{deepmind2026gemini31flashlite} & 33.7\% & 48.1\% & 45.7\% & 30.1\% & 47.9\% & 46.0\% & 47.6\% & 9.7\% & 21.1\% & 35.9\% & 9.3\% & 20.8\% & 12.4\% & 5.5\% \\
Gemini 3 Flash\cite{deepmind2025gemini3flash} & \cellcolor{secondblue}{47.6\%} & 33.6\% & 33.8\% & 45.9\% & 37.5\% & 38.5\% & 56.2\% & 5.0\% & 13.5\% & 51.3\% & 5.1\% & 12.2\% & 24.9\% & 2.0\% \\
GPT-6 Astra\textsuperscript{T}\cite{openai2026gpt6astra} & 35.1\% & 27.8\% & 31.2\% & 43.0\% & 27.8\% & 32.7\% & 57.2\% & 16.1\% & 24.3\% & 55.4\% & 14.9\% & 24.3\% & 21.5\% & 6.9\% \\
GPT-5.6 Sol\textsuperscript{T}\cite{openai2026gpt56sol} & 34.1\% & 23.9\% & 28.8\% & 42.1\% & 24.6\% & 30.1\% & 55.7\% & 12.9\% & 23.0\% & 55.8\% & 12.5\% & 22.4\% & 20.5\% & 5.6\% \\
GPT-5.6 Terra\textsuperscript{T}\cite{openai2026gpt56sol} & 33.7\% & 24.8\% & 28.9\% & 40.5\% & 27.1\% & 31.0\% & 58.9\% & 14.5\% & 23.4\% & 60.2\% & 12.9\% & 22.5\% & 22.3\% & 6.3\% \\
GPT-5.5\cite{openai2026gpt55} & 38.0\% & 19.4\% & 26.2\% & 53.4\% & 17.5\% & 26.3\% & \cellcolor{bestblue}{65.4\%} & 17.6\% & 25.1\% & 53.6\% & 17.9\% & 24.7\% & 27.1\% & 5.8\% \\
GPT-5.4\cite{openai2026gpt54} & 37.9\% & 12.2\% & 24.1\% & 37.6\% & 13.9\% & 25.9\% & 37.8\% & 12.9\% & 24.4\% & 36.0\% & 12.0\% & 23.5\% & 14.0\% & 3.2\% \\
GPT-5.3\cite{openai2026gpt53} & 35.2\% & 11.9\% & 23.5\% & 37.3\% & 13.9\% & 26.7\% & 56.5\% & 14.9\% & 22.5\% & 62.3\% & 15.4\% & 22.5\% & 21.1\% & 2.9\% \\
MiMo-V2-Omni\cite{mimo_v2_omni} & 28.7\% & 40.7\% & 36.5\% & 21.3\% & 36.9\% & 33.4\% & 36.5\% & 8.8\% & 17.6\% & 38.2\% & 9.2\% & 17.3\% & 9.4\% & 4.6\% \\
Muse Spark 1.3 Contributor\textsuperscript{T}\cite{meta2026musespark13} & 38.7\% & 20.5\% & 26.7\% & 48.6\% & 15.0\% & 21.2\% & \cellcolor{secondblue}{64.2\%} & 14.8\% & 25.5\% & \cellcolor{secondblue}{64.4\%} & 14.9\% & 24.9\% & \cellcolor{secondblue}{28.1\%} & 4.5\% \\
Qwen3.7-Plus\cite{qwen37plus} & 33.3\% & 18.4\% & 28.0\% & 34.6\% & 21.4\% & 30.3\% & 55.4\% & 18.4\% & 26.7\% & 43.4\% & 18.8\% & 27.0\% & 15.7\% & 6.5\% \\
Qwen3.6-Plus\cite{qwen36plus} & 32.8\% & 16.6\% & 27.1\% & 33.5\% & 14.7\% & 26.1\% & 48.4\% & 7.4\% & 19.0\% & 42.5\% & 6.0\% & 19.9\% & 13.7\% & 1.6\% \\
Qwen3.6-Flash\cite{qwen36flash} & 25.1\% & 24.6\% & 32.0\% & 28.7\% & 19.2\% & 29.8\% & 62.7\% & 19.4\% & 27.1\% & 56.7\% & 18.6\% & 25.8\% & 16.0\% & 6.0\% \\
Qwen3.5-Plus\cite{qwen3.5} & 28.6\% & 21.4\% & 27.1\% & 29.0\% & 12.3\% & 24.1\% & 47.6\% & 9.7\% & 19.3\% & 48.9\% & 8.3\% & 20.0\% & 12.7\% & 1.6\% \\
Qwen3.5-Flash\cite{qwen3.5} & 27.8\% & 21.6\% & 29.8\% & 28.9\% & 15.8\% & 23.1\% & 52.4\% & 8.8\% & 19.4\% & 54.5\% & 8.9\% & 20.1\% & 15.0\% & 2.3\% \\
\midrule
\rowcolor{gray!15} \multicolumn{15}{l}{\textit{\textbf{Open-Weight Models}}} \\
GLM-5.3-Flash\textsuperscript{T}\cite{zai2026glm53flash} & 32.4\% & 38.4\% & 38.1\% & 35.2\% & \cellcolor{bestblue}{45.4\%} & \cellcolor{bestblue}{43.8\%} & \cellcolor{bestblue}{63.6\%} & \cellcolor{bestblue}{32.7\%} & \cellcolor{bestblue}{33.5\%} & \cellcolor{secondblue}{57.4\%} & \cellcolor{bestblue}{33.0\%} & \cellcolor{bestblue}{34.1\%} & \cellcolor{bestblue}{19.9\%} & \cellcolor{bestblue}{20.6\%} \\
GLM-4.6V\cite{glm_46v} & 30.8\% & \cellcolor{bestblue}{47.5\%} & \cellcolor{bestblue}{43.0\%} & 27.1\% & \cellcolor{secondblue}{44.9\%} & \cellcolor{secondblue}{42.3\%} & 29.2\% & 9.5\% & 20.0\% & 44.2\% & 10.2\% & 20.6\% & 9.9\% & 5.3\% \\
GLM-4.5V\cite{zai2025glm45v} & 31.9\% & 31.6\% & 32.1\% & 29.0\% & 23.4\% & 26.3\% & 19.9\% & 15.9\% & 22.5\% & 39.6\% & 17.8\% & 24.2\% & 8.8\% & 7.7\% \\
Intern-S1-Pro\cite{zou2026interns1proscientificmultimodalfoundation} & 29.4\% & 17.4\% & 26.8\% & 26.2\% & 10.5\% & 22.6\% & 27.8\% & 2.7\% & 13.7\% & 30.3\% & 3.2\% & 14.0\% & 7.3\% & 0.4\% \\
InternVL 3.5-241B-A28B\cite{wang2025internvl3_5} & 32.9\% & 31.3\% & 34.1\% & 26.1\% & 17.5\% & 24.8\% & 50.1\% & 5.4\% & 16.2\% & 41.8\% & 5.2\% & 15.3\% & 13.0\% & 1.7\% \\
InternVL 3.5-30B-A3B\cite{wang2025internvl3_5} & 27.3\% & 4.9\% & 20.0\% & 24.0\% & 7.0\% & 15.3\% & 35.4\% & 0.6\% & 7.6\% & 41.3\% & 1.3\% & 7.3\% & 9.1\% & 0.0\% \\
InternVL 3.5-14B\cite{wang2025internvl3_5} & 22.8\% & 7.6\% & 22.2\% & 31.9\% & 7.5\% & 22.5\% & 28.5\% & 4.7\% & 12.9\% & 41.3\% & 5.0\% & 11.9\% & 8.7\% & 0.6\% \\
InternVL 3.5-8B\cite{wang2025internvl3_5} & 28.5\% & 19.7\% & 34.2\% & 24.5\% & 9.5\% & 29.1\% & 24.7\% & 4.2\% & 12.3\% & 30.8\% & 2.5\% & 11.4\% & 7.2\% & 0.6\% \\
InternVL 3.5-4B\cite{wang2025internvl3_5} & 25.6\% & 3.1\% & 19.8\% & 26.4\% & 12.0\% & 23.7\% & 22.4\% & 1.0\% & 8.1\% & 36.7\% & 1.0\% & 7.5\% & 7.1\% & 0.0\% \\
Kimi K2.6\cite{kimi_k26} & \cellcolor{bestblue}{36.1\%} & 22.1\% & 28.5\% & 32.3\% & 12.2\% & 20.8\% & 42.8\% & 12.3\% & 21.9\% & 28.0\% & 10.8\% & 20.4\% & 11.3\% & 3.7\% \\
Kimi K2.5\cite{team2026kimi} & \cellcolor{secondblue}{34.3\%} & 27.1\% & 30.4\% & 32.7\% & 18.5\% & 23.9\% & 39.4\% & 5.0\% & 15.6\% & 30.1\% & 4.7\% & 14.4\% & 10.7\% & 1.1\% \\
MiMo-V2.5\cite{mimo_v25} & 33.1\% & \cellcolor{secondblue}{43.3\%} & 38.3\% & 27.3\% & 43.5\% & 40.1\% & 51.2\% & \cellcolor{secondblue}{26.6\%} & \cellcolor{secondblue}{30.7\%} & 49.7\% & \cellcolor{secondblue}{27.8\%} & \cellcolor{secondblue}{31.5\%} & 15.3\% & \cellcolor{secondblue}{19.0\%} \\
Qwen3.8-27B\cite{qwen2026qwen38_27b} & 26.2\% & 34.7\% & 35.2\% & 28.8\% & 17.0\% & 26.1\% & \cellcolor{secondblue}{61.9\%} & 22.8\% & 27.6\% & 49.4\% & 21.9\% & 28.1\% & 15.3\% & 8.5\% \\
Qwen3.6-27B\cite{qwen2026qwen36_27b} & 30.8\% & 36.0\% & 35.2\% & 36.4\% & 13.2\% & 22.8\% & 60.4\% & 21.7\% & 27.8\% & 53.4\% & 19.5\% & 27.3\% & \cellcolor{secondblue}{18.4\%} & 6.2\% \\
Qwen3.5-397B-A17B\cite{qwen3.5} & 27.4\% & 3.3\% & 12.6\% & 35.8\% & 13.5\% & 23.5\% & 47.9\% & 2.8\% & 13.0\% & 46.2\% & 4.4\% & 13.9\% & 14.1\% & 0.2\% \\
Qwen3.5-122B-A10B\cite{qwen3.5} & 32.6\% & 29.1\% & 31.6\% & \cellcolor{bestblue}{37.8\%} & 26.0\% & 31.6\% & 49.7\% & 6.1\% & 19.4\% & 49.0\% & 6.8\% & 18.8\% & 15.4\% & 2.1\% \\
Qwen3.5-35B-A3B\cite{qwen3.5} & 29.5\% & 27.6\% & 30.5\% & 27.4\% & 0.6\% & 0.8\% & 53.2\% & 6.1\% & 14.5\% & \cellcolor{bestblue}{57.5\%} & 7.2\% & 18.5\% & 15.5\% & 0.0\% \\
Qwen3.5-27B\cite{qwen3.5} & 31.6\% & 42.8\% & \cellcolor{secondblue}{39.9\%} & \cellcolor{secondblue}{37.1\%} & 22.8\% & 28.5\% & 57.8\% & 7.3\% & 20.4\% & 50.6\% & 7.5\% & 21.0\% & 17.8\% & 2.9\% \\
Qwen3.5-9B\cite{qwen3.5} & 29.0\% & 15.9\% & 25.8\% & 29.0\% & 22.9\% & 32.5\% & 53.2\% & 10.3\% & 20.9\% & 52.2\% & 11.5\% & 22.1\% & 15.1\% & 2.8\% \\
Qwen3.5-4B\cite{qwen3.5} & 30.9\% & 12.9\% & 23.2\% & 30.2\% & 12.4\% & 23.1\% & 47.5\% & 6.8\% & 17.8\% & 47.5\% & 6.3\% & 18.0\% & 15.4\% & 1.2\% \\
\midrule
\rowcolor{gray!15} \multicolumn{15}{l}{\textit{\textbf{Spatially Specialized Models}}} \\
SpaceOm\cite{SpaceOm2025} & \cellcolor{bestblue}{24.7\%} & \cellcolor{secondblue}{2.8\%} & \cellcolor{secondblue}{6.7\%} & \cellcolor{secondblue}{24.5\%} & \cellcolor{bestblue}{4.2\%} & \cellcolor{bestblue}{11.4\%} & 35.2\% & \cellcolor{bestblue}{2.4\%} & \cellcolor{bestblue}{8.6\%} & 32.8\% & \cellcolor{secondblue}{1.8\%} & \cellcolor{secondblue}{6.5\%} & 8.0\% & \cellcolor{secondblue}{0.1\%} \\
SpaceR\cite{ouyang2025spacer} & 20.6\% & 2.5\% & 6.3\% & 24.0\% & 1.1\% & 1.8\% & \cellcolor{secondblue}{42.4\%} & \cellcolor{secondblue}{2.1\%} & \cellcolor{secondblue}{6.5\%} & \cellcolor{secondblue}{48.8\%} & \cellcolor{bestblue}{2.3\%} & \cellcolor{bestblue}{6.9\%} & \cellcolor{secondblue}{9.4\%} & 0.0\% \\
SpaceThinker\cite{SpaceThinker2025} & \cellcolor{secondblue}{22.6\%} & 1.2\% & 3.0\% & \cellcolor{bestblue}{25.4\%} & 1.8\% & \cellcolor{secondblue}{5.9\%} & 32.8\% & 0.6\% & 5.1\% & 34.3\% & 0.7\% & 4.7\% & 7.9\% & 0.0\% \\
ViLaSR\cite{wu2025reinforcing} & 18.8\% & \cellcolor{bestblue}{4.3\%} & \cellcolor{bestblue}{7.6\%} & 22.1\% & \cellcolor{secondblue}{2.0\%} & 3.5\% & \cellcolor{bestblue}{43.1\%} & 1.9\% & 3.5\% & \cellcolor{bestblue}{49.7\%} & 1.5\% & 3.2\% & \cellcolor{bestblue}{9.7\%} & \cellcolor{bestblue}{0.3\%} \\
\midrule
\rowcolor{gray!15} \multicolumn{15}{l}{\textit{\textbf{Human Baseline}}} \\
Human & 85.8\% & 77.9\% & 57.8\% & 94.6\% & 75.4\% & 59.5\% & 82.5\% & 53.8\% & 47.2\% & 82.5\% & 72.5\% & 57.5\% & -- & -- \\
\bottomrule
\end{tabular}
\end{table*}

\definecolor{bestblue}{HTML}{BBEBFF}
\definecolor{secondblue}{HTML}{E6F5FF}

\begin{table*}[t]
\centering
\caption{Video-task results (\%). Dual columns pair behavior and visibility questions from one trajectory. \textsuperscript{T}: thinking-only endpoint at low effort (minimum effort available for Muse Spark). \colorbox{bestblue}{Dark} and \colorbox{secondblue}{light} blue mark the best and second-best model within each group.}
\label{tab:results_video}
\scriptsize
\setlength{\tabcolsep}{2.2pt}
\begin{tabular}{l|ccc|ccc|ccc|cc}
\toprule
\multirow{3}{*}{\textbf{Model}} & \multicolumn{6}{c|}{\textbf{Flight Behavior Recognition}} & \multicolumn{3}{c|}{\textbf{Landmark Visibility}} & \multicolumn{2}{c}{\textbf{Coordination}} \\
\cmidrule(lr){2-7} \cmidrule(lr){8-10} \cmidrule(lr){11-12}
& \multicolumn{3}{c|}{\textbf{Composite Level}} & \multicolumn{3}{c|}{\textbf{Atomic Level}} & \multicolumn{3}{c|}{\textbf{Counting}} & \multirow{2}{*}{Dual Sem.} & \multirow{2}{*}{Dual Evi.} \\
& Acc & F1@0.5 & mtIoU  & F1 & F1@0.5 & mtIoU  & Acc & F1@0.5 & mtIoU & & \\
\midrule
\rowcolor{gray!15} \multicolumn{12}{l}{\textit{\textbf{Proprietary Models}}} \\
Gemini 3.8 Flash\textsuperscript{T}\cite{deepmind2026gemini38flash} & \cellcolor{bestblue}{56.4\%} & 41.4\% & 43.0\% & \cellcolor{secondblue}{38.8\%} & 17.2\% & 14.6\% & 46.0\% & 25.3\% & 29.0\% & 6.9\% & 8.2\% \\
Gemini 3.1 Pro Preview\textsuperscript{T}\cite{deepmind2026gemini31pro} & \cellcolor{secondblue}{54.8\%} & 40.2\% & 37.9\% & 32.9\% & 9.0\% & 8.6\% & 33.5\% & 20.4\% & 30.3\% & 1.2\% & 2.1\% \\
Gemini 3.1 Flash Lite\cite{deepmind2026gemini31flashlite} & 23.7\% & 23.0\% & 26.0\% & 19.7\% & 11.6\% & 10.3\% & 52.8\% & 44.1\% & 42.2\% & 7.3\% & 5.5\% \\
Gemini 3 Flash\cite{deepmind2025gemini3flash} & 48.3\% & \cellcolor{secondblue}{47.7\%} & \cellcolor{secondblue}{52.4\%} & 35.0\% & \cellcolor{bestblue}{27.5\%} & \cellcolor{secondblue}{22.0\%} & \cellcolor{secondblue}{55.2\%} & 45.0\% & 44.3\% & 11.4\% & \cellcolor{secondblue}{12.6\%} \\
MiMo-V2-Omni\cite{mimo_v2_omni} & 26.9\% & 25.0\% & 26.9\% & 30.1\% & 21.3\% & 17.4\% & \cellcolor{bestblue}{59.4\%} & 52.7\% & \cellcolor{bestblue}{53.7\%} & 10.6\% & 11.5\% \\
Muse Spark 1.3 Contributor\textsuperscript{T}\cite{meta2026musespark13} & 49.2\% & \cellcolor{bestblue}{50.0\%} & \cellcolor{bestblue}{53.1\%} & \cellcolor{bestblue}{39.2\%} & \cellcolor{secondblue}{27.2\%} & \cellcolor{bestblue}{22.3\%} & 47.9\% & 48.0\% & 44.2\% & 12.6\% & \cellcolor{bestblue}{16.2\%} \\
Qwen3.7-Plus\cite{qwen37plus} & 6.5\% & 6.2\% & 6.4\% & 33.8\% & 24.2\% & 18.3\% & 53.6\% & \cellcolor{secondblue}{56.8\%} & 49.2\% & 14.0\% & 11.5\% \\
Qwen3.6-Plus\cite{qwen36plus} & 9.8\% & 9.3\% & 9.6\% & 30.1\% & 15.6\% & 15.3\% & 50.8\% & 52.2\% & 46.1\% & \cellcolor{bestblue}{14.8\%} & 8.5\% \\
Qwen3.6-Flash\cite{qwen36flash} & 32.3\% & 28.1\% & 32.0\% & 24.9\% & 16.9\% & 12.9\% & 52.2\% & 52.7\% & 46.7\% & 11.5\% & 7.3\% \\
Qwen3.5-Plus\cite{qwen3.5} & 2.4\% & 2.4\% & 2.5\% & 32.1\% & 17.5\% & 17.0\% & 53.2\% & 54.2\% & 46.8\% & \cellcolor{secondblue}{14.4\%} & 9.1\% \\
Qwen3.5-Flash\cite{qwen3.5} & 32.6\% & 29.8\% & 33.1\% & 26.9\% & 17.8\% & 13.9\% & 52.3\% & \cellcolor{bestblue}{58.3\%} & \cellcolor{secondblue}{51.3\%} & 9.3\% & 9.5\% \\
\midrule
\rowcolor{gray!15} \multicolumn{12}{l}{\textit{\textbf{Open-Weight Models}}} \\
GLM-5.3-Flash\textsuperscript{T}\cite{zai2026glm53flash} & 26.0\% & \cellcolor{secondblue}{24.5\%} & \cellcolor{secondblue}{25.8\%} & 30.3\% & \cellcolor{secondblue}{21.2\%} & \cellcolor{bestblue}{18.0\%} & 49.1\% & 43.9\% & 41.4\% & 11.8\% & 9.4\% \\
GLM-4.6V\cite{glm_46v} & \cellcolor{secondblue}{31.2\%} & 1.2\% & 6.0\% & \cellcolor{bestblue}{32.2\%} & 7.5\% & 6.2\% & 31.2\% & 17.4\% & 33.8\% & 0.1\% & 0.5\% \\
InternVL 3.5-30B-A3B\cite{wang2025internvl3_5} & 13.3\% & 1.9\% & 4.3\% & 24.5\% & 5.8\% & 5.4\% & 43.9\% & 8.2\% & 12.6\% & 7.2\% & 0.3\% \\
InternVL 3.5-14B\cite{wang2025internvl3_5} & 23.4\% & 2.3\% & 7.3\% & 21.8\% & 3.9\% & 3.8\% & \cellcolor{secondblue}{54.5\%} & 14.4\% & 22.7\% & 3.0\% & 0.9\% \\
InternVL 3.5-8B\cite{wang2025internvl3_5} & 22.4\% & 6.2\% & 10.6\% & 22.9\% & 4.2\% & 4.6\% & 51.9\% & 16.7\% & 21.3\% & 6.4\% & 0.3\% \\
InternVL 3.5-4B\cite{wang2025internvl3_5} & 25.8\% & 2.3\% & 5.8\% & 20.7\% & 4.3\% & 3.4\% & 34.6\% & 17.1\% & 32.3\% & 0.4\% & 0.5\% \\
Kimi K2.6\cite{kimi_k26} & 11.9\% & 10.7\% & 11.1\% & \cellcolor{secondblue}{31.4\%} & 20.0\% & 16.7\% & 39.9\% & 48.3\% & 43.3\% & 8.5\% & 7.8\% \\
Kimi K2.5\cite{team2026kimi} & 13.3\% & 12.4\% & 13.2\% & 30.0\% & \cellcolor{secondblue}{21.2\%} & 17.6\% & 50.5\% & 50.3\% & 45.8\% & 11.5\% & \cellcolor{bestblue}{10.2\%} \\
MiMo-V2.5\cite{mimo_v25} & \cellcolor{bestblue}{35.6\%} & \cellcolor{bestblue}{33.4\%} & \cellcolor{bestblue}{36.9\%} & 29.5\% & \cellcolor{bestblue}{21.4\%} & \cellcolor{bestblue}{18.0\%} & \cellcolor{bestblue}{58.5\%} & \cellcolor{bestblue}{54.4\%} & \cellcolor{secondblue}{50.4\%} & \cellcolor{secondblue}{11.9\%} & \cellcolor{secondblue}{10.0\%} \\
Qwen3.8-27B\cite{qwen2026qwen38_27b} & 15.3\% & 6.8\% & 7.5\% & 27.8\% & 12.4\% & 8.6\% & 43.2\% & 16.3\% & 23.6\% & 8.9\% & 0.5\% \\
Qwen3.6-27B\cite{qwen2026qwen36_27b} & 19.9\% & 12.8\% & 14.1\% & 29.1\% & 18.9\% & 15.8\% & 52.7\% & 43.3\% & 42.3\% & \cellcolor{bestblue}{13.4\%} & 7.7\% \\
Qwen3.5-397B-A17B\cite{qwen3.5} & 5.2\% & 1.0\% & 1.5\% & 26.6\% & 10.3\% & 8.4\% & 40.4\% & 21.2\% & 31.8\% & 5.7\% & 1.9\% \\
Qwen3.5-122B-A10B\cite{qwen3.5} & 21.0\% & 17.8\% & 19.9\% & 23.4\% & 13.3\% & 11.9\% & 48.4\% & 46.7\% & 49.7\% & 5.4\% & 6.5\% \\
Qwen3.5-35B-A3B\cite{qwen3.5} & 21.0\% & 15.7\% & 16.7\% & 27.7\% & 17.8\% & 16.0\% & 54.1\% & 50.0\% & 48.6\% & 8.9\% & 9.0\% \\
Qwen3.5-27B\cite{qwen3.5} & 14.6\% & 13.4\% & 14.3\% & 29.7\% & 20.1\% & \cellcolor{secondblue}{17.7\%} & 51.4\% & \cellcolor{secondblue}{52.2\%} & 48.7\% & 11.2\% & 9.7\% \\
Qwen3.5-9B\cite{qwen3.5} & 9.5\% & 7.3\% & 8.3\% & 28.2\% & 17.8\% & 16.0\% & 50.7\% & 33.8\% & 36.8\% & 8.0\% & 6.0\% \\
Qwen3.5-4B\cite{qwen3.5} & 6.4\% & 1.1\% & 1.1\% & 10.0\% & 0.9\% & 0.8\% & 33.9\% & 45.2\% & \cellcolor{bestblue}{52.0\%} & 0.6\% & 0.4\% \\
\midrule
\rowcolor{gray!15} \multicolumn{12}{l}{\textit{\textbf{Spatially Specialized Models}}} \\
SpaceOm\cite{SpaceOm2025} & \cellcolor{bestblue}{26.5\%} & 0.0\% & \cellcolor{bestblue}{2.5\%} & \cellcolor{secondblue}{25.9\%} & \cellcolor{bestblue}{4.0\%} & \cellcolor{secondblue}{3.1\%} & 32.3\% & \cellcolor{bestblue}{17.6\%} & \cellcolor{bestblue}{34.2\%} & 0.3\% & 0.0\% \\
SpaceR\cite{ouyang2025spacer} & \cellcolor{secondblue}{21.4\%} & 0.0\% & \cellcolor{secondblue}{1.9\%} & \cellcolor{bestblue}{26.2\%} & \cellcolor{bestblue}{4.0\%} & \cellcolor{bestblue}{3.5\%} & \cellcolor{secondblue}{33.7\%} & \cellcolor{secondblue}{16.5\%} & \cellcolor{secondblue}{30.8\%} & 0.4\% & 0.0\% \\
SpaceThinker\cite{SpaceThinker2025} & 17.2\% & 0.0\% & 1.3\% & 23.1\% & \cellcolor{secondblue}{3.7\%} & 2.6\% & \cellcolor{secondblue}{33.7\%} & 14.3\% & 27.1\% & \cellcolor{secondblue}{0.9\%} & 0.0\% \\
ViLaSR\cite{wu2025reinforcing} & 5.2\% & 0.0\% & 0.4\% & 9.9\% & 0.4\% & 1.1\% & \cellcolor{bestblue}{35.9\%} & 10.4\% & 20.6\% & \cellcolor{bestblue}{1.6\%} & 0.0\% \\
\midrule
\rowcolor{gray!15} \multicolumn{12}{l}{\textit{\textbf{Human Baseline}}} \\
Human & 83.9\% & 62.9\% & 53.3\% & 76.3\% & 83.2\% & 61.6\% & 95.0\% & 96.7\% & 83.7\% & -- & -- \\
\bottomrule
\end{tabular}
\end{table*}

\subsection{Experimental Setup}
\noindent \ding{224} \textbf{Model selection.}
We report 45 models with valid image-task results: 21 \textit{proprietary}, 20 \textit{open-weight}, and 4 \textit{spatially specialized} models. Among them, 32 support video input and have valid video results.

\noindent \ding{224} \textbf{Evaluation settings.}
We use \textit{Instant} mode to ensure latency stability when supported. Models available only through reasoning-enabled endpoints are marked \textsuperscript{T} and evaluated at \textit{low} reasoning effort (\textit{minimum} available for Muse Spark). A unified compatible parser audits all responses; calling settings, prompts, coverage, and parser rules are provided on the Evaluation page.

\subsection{Evaluation Metrics}
Structured outputs separate semantic decisions from grounding evidence. Images use option accuracy, BBox Acc@0.5 ($\text{IoU}\ge0.5$), and mIoU. Videos use composite exact-set accuracy, atomic label-set F1, visibility-count accuracy, and class-aware one-to-one temporal F1@0.5 and mtIoU. For an aligned Agent-State and World-State pair $i$, let $s_i^a,s_i^w$ denote semantic correctness and $g_i^a,g_i^w$ denote task-specific grounding success at the 0.5 threshold. We define
\begin{equation}
\begin{aligned}
A_{\mathrm{dual}}^{\mathrm{sem}}&=\frac{1}{N}\sum_i s_i^as_i^w,\\
A_{\mathrm{dual}}^{\mathrm{evi}}&=\frac{1}{N}\sum_i g_i^ag_i^w.
\end{aligned}
\end{equation}
\textbf{Dual Semantic Acc} and \textbf{Dual Evidence Acc} independently measure paired semantic and grounding co-success. Image records pair relative-position and action--observation tasks within the same scene--Agent--Landmark--difficulty stratum and harmonically average both pair scores; video records pair behavior and visibility questions from one trajectory. Full metric, matching, pairing, and candidate-domain rules are provided on the Evaluation page.

\subsection{Main Results}
\noindent \ding{96} \textbf{Image tasks.}
Table~\ref{tab:results_image} shows that World-State semantics often exceed Agent-State semantics, while grounding follows a different ranking. Gemini~3.8 Flash\textsuperscript{T} substantially raises the frontier, leading both Agent-State tasks (60.1\% and 74.9\% Acc) and most grounding columns; GPT-5.5 still leads Self-relative Position accuracy (65.4\%). Muse Spark\textsuperscript{T} reaches 64.2\% and 64.4\% on the two World-State tasks, but only 14.8\% and 14.9\% BBox Acc@0.5. Among open-weight models, GLM-5.3-Flash\textsuperscript{T} leads both World-State grounding columns (32.7\% and 33.0\% BBox Acc@0.5). Spatially specialized models show no consistent advantage. Gemini~3.8 Flash\textsuperscript{T} consequently leads both Dual Semantic and Dual Evidence Acc (44.4\% and 34.3\%), yet both remain well below its strongest marginal scores.

\noindent \ding{96} \textbf{Video tasks.}
Table~\ref{tab:results_video} reveals uneven behavior understanding and temporal localization. Visibility counting is generally stronger than behavior recognition, while Qwen3.7-Plus's 33.8\% atomic F1 but 6.5\% composite accuracy exposes a behavior-hierarchy gap. Gemini~3.8 Flash\textsuperscript{T} leads composite accuracy (56.4\%), whereas Muse Spark\textsuperscript{T} leads composite temporal F1@0.5 and mtIoU (50.0\% and 53.1\%); MiMo-V2-Omni leads visibility accuracy and mtIoU (59.4\% and 53.7\%). Under paired evaluation, Qwen3.6-Plus leads Dual Semantic Acc (14.8\%), while Muse Spark\textsuperscript{T} leads Dual Evidence Acc (16.2\%), again separating joint state recognition from joint temporal grounding.

\subsection{Analysis and Discussion}
\begin{table}[!t]
\centering
\caption{Top-5 Dual Semantic/Evidence Acc (\%). Models are ranked within each group by Overall Sem. Pos., Act.--Obs., and Beh.--Vis. denote the three task pairs. Dark/light blue mark the best/second-best in each group; \textsuperscript{T}: thinking-only.}
\label{tab:dual_cognition_overall}
\setlength{\tabcolsep}{1.2pt}
\renewcommand{\arraystretch}{0.90}
\resizebox{\columnwidth}{!}{%
\begin{tabular}{l|cc|cc|cc|cc}
\toprule
\multirow{2}{*}{\textbf{Model}} & \multicolumn{2}{c|}{\textbf{Pos.}} & \multicolumn{2}{c|}{\textbf{Act.--Obs.}} & \multicolumn{2}{c|}{\textbf{Beh.--Vis.}} & \multicolumn{2}{c}{\textbf{Overall}} \\
\cmidrule(lr){2-3} \cmidrule(lr){4-5} \cmidrule(lr){6-7} \cmidrule(lr){8-9}
& Sem. & Evi. & Sem. & Evi. & Sem. & Evi. & Sem. & Evi. \\
\midrule
\rowcolor{gray!15} \multicolumn{9}{l}{\textit{\textbf{Proprietary Models}}} \\
Muse Spark 1.3 Contributor\textsuperscript{T} & \cellcolor{secondblue}{24.7\%} & \cellcolor{bestblue}{6.7\%} & \cellcolor{bestblue}{32.6\%} & \cellcolor{secondblue}{3.4\%} & 12.6\% & \cellcolor{bestblue}{16.2\%} & \cellcolor{bestblue}{17.4\%} & \cellcolor{secondblue}{7.1\%} \\
Gemini 3 Flash & \cellcolor{bestblue}{25.6\%} & 1.9\% & \cellcolor{secondblue}{24.3\%} & 2.1\% & 11.4\% & \cellcolor{secondblue}{12.6\%} & \cellcolor{secondblue}{15.7\%} & 3.4\% \\
Qwen3.7-Plus & 16.4\% & \cellcolor{secondblue}{6.5\%} & 15.1\% & \cellcolor{bestblue}{6.4\%} & 14.0\% & 11.5\% & 14.8\% & \cellcolor{bestblue}{8.3\%} \\
Qwen3.6-Plus & 13.5\% & 1.7\% & 14.0\% & 1.5\% & \cellcolor{bestblue}{14.8\%} & 8.5\% & 14.3\% & 2.6\% \\
Qwen3.5-Plus & 11.2\% & 3.0\% & 14.6\% & 1.1\% & \cellcolor{secondblue}{14.4\%} & 9.1\% & 13.5\% & 2.7\% \\
\midrule
\rowcolor{gray!15} \multicolumn{9}{l}{\textit{\textbf{Open-Weight Models}}} \\
Qwen3.6-27B & \cellcolor{secondblue}{17.5\%} & 13.0\% & \cellcolor{secondblue}{19.3\%} & 4.1\% & \cellcolor{bestblue}{13.4\%} & 7.7\% & \cellcolor{bestblue}{15.5\%} & 6.9\% \\
GLM-5.3-Flash\textsuperscript{T} & \cellcolor{bestblue}{19.3\%} & \cellcolor{bestblue}{20.7\%} & \cellcolor{bestblue}{20.5\%} & \cellcolor{bestblue}{20.5\%} & 11.8\% & 9.4\% & \cellcolor{secondblue}{14.8\%} & \cellcolor{secondblue}{12.9\%} \\
Qwen3.5-27B & 17.4\% & 3.8\% & 18.2\% & 2.3\% & 11.2\% & \cellcolor{secondblue}{9.7\%} & 13.8\% & 4.5\% \\
MiMo-V2.5 & 16.2\% & \cellcolor{secondblue}{18.9\%} & 14.6\% & \cellcolor{secondblue}{19.1\%} & \cellcolor{secondblue}{11.9\%} & \cellcolor{bestblue}{10.0\%} & 13.4\% & \cellcolor{bestblue}{13.1\%} \\
Qwen3.5-35B-A3B & 15.3\% & 2.4\% & 15.7\% & 0.0\% & 8.9\% & 9.0\% & 11.3\% & 0.0\% \\
\midrule
\rowcolor{gray!15} \multicolumn{9}{l}{\textit{\textbf{Spatially Specialized Models}}} \\
ViLaSR & \cellcolor{bestblue}{9.6\%} & \cellcolor{bestblue}{0.5\%} & \cellcolor{secondblue}{9.8\%} & \cellcolor{bestblue}{0.2\%} & \cellcolor{bestblue}{1.6\%} & 0.0\% & \cellcolor{bestblue}{2.7\%} & 0.0\% \\
SpaceThinker & 7.5\% & 0.0\% & 8.3\% & 0.0\% & \cellcolor{secondblue}{0.9\%} & 0.0\% & \cellcolor{secondblue}{1.6\%} & 0.0\% \\
SpaceR & 8.1\% & \cellcolor{secondblue}{0.2\%} & \cellcolor{bestblue}{11.2\%} & 0.0\% & 0.4\% & 0.0\% & 0.8\% & 0.0\% \\
SpaceOm & \cellcolor{secondblue}{8.4\%} & \cellcolor{secondblue}{0.2\%} & 7.6\% & \cellcolor{secondblue}{0.1\%} & 0.3\% & 0.0\% & 0.6\% & 0.0\% \\
\bottomrule
\end{tabular}%
}
\end{table}

\begin{table}[t]
\centering
\caption{Top-5 Dual Semantic/Evidence Acc (\%) on the real-scene 3DGS pilot (16 models). Models are ranked within each group by Overall Sem. Dark/light blue mark the best/second-best in each group; \textsuperscript{T}: thinking-only.}
\label{tab:3dgs_dual_cognition}
\scriptsize
\setlength{\tabcolsep}{1.7pt}
\begin{tabular*}{\columnwidth}{@{\extracolsep{\fill}}l|cc|cc|cc@{}}
\toprule
\multirow{2}{*}{Model} & \multicolumn{2}{c|}{Pos.} & \multicolumn{2}{c|}{Act.--Obs.} & \multicolumn{2}{c}{Overall} \\
& Sem. & Evi. & Sem. & Evi. & Sem. & Evi. \\
\midrule
\rowcolor{gray!15} \multicolumn{7}{l}{\textit{\textbf{Proprietary Models}}} \\
Gemini 3.8 Flash\textsuperscript{T} & \cellcolor{bestblue}{15.2\%} & \cellcolor{bestblue}{66.7\%} & \cellcolor{bestblue}{13.6\%} & \cellcolor{bestblue}{72.7\%} & \cellcolor{bestblue}{14.4\%} & \cellcolor{bestblue}{69.6\%} \\
Muse Spark 1.3 Contributor\textsuperscript{T} & \cellcolor{secondblue}{13.6\%} & 10.6\% & \cellcolor{secondblue}{10.6\%} & 6.1\% & \cellcolor{secondblue}{11.9\%} & 7.7\% \\
Gemini 3 Flash & \cellcolor{secondblue}{13.6\%} & \cellcolor{secondblue}{50.0\%} & \cellcolor{secondblue}{10.6\%} & \cellcolor{secondblue}{53.0\%} & \cellcolor{secondblue}{11.9\%} & \cellcolor{secondblue}{51.5\%} \\
GPT-5.6 Sol\textsuperscript{T} & \cellcolor{bestblue}{15.2\%} & 6.1\% & 9.1\% & 6.1\% & 11.4\% & 6.1\% \\
GPT-5.5 & \cellcolor{secondblue}{13.6\%} & 4.5\% & 7.6\% & 1.5\% & 9.7\% & 2.3\% \\
\midrule
\rowcolor{gray!15} \multicolumn{7}{l}{\textit{\textbf{Open-Weight Models}}} \\
GLM-5.3-Flash\textsuperscript{T} & 13.6\% & \cellcolor{secondblue}{28.8\%} & \cellcolor{bestblue}{13.6\%} & \cellcolor{bestblue}{24.2\%} & \cellcolor{bestblue}{13.6\%} & \cellcolor{bestblue}{26.3\%} \\
MiMo-V2.5 & \cellcolor{bestblue}{16.7\%} & \cellcolor{bestblue}{30.3\%} & \cellcolor{secondblue}{10.6\%} & \cellcolor{secondblue}{21.2\%} & \cellcolor{secondblue}{13.0\%} & \cellcolor{secondblue}{25.0\%} \\
Qwen3.6-27B & \cellcolor{secondblue}{15.2\%} & \cellcolor{secondblue}{28.8\%} & \cellcolor{secondblue}{10.6\%} & 7.6\% & 12.5\% & 12.0\% \\
Kimi K2.6 & 10.6\% & 21.2\% & \cellcolor{bestblue}{13.6\%} & 3.0\% & 11.9\% & 5.3\% \\
Kimi K2.5 & \cellcolor{secondblue}{15.2\%} & 21.2\% & 9.1\% & 4.5\% & 11.4\% & 7.5\% \\
\bottomrule
\end{tabular*}
\end{table}

\noindent \ding{47} \textbf{Dual-cognition coordination.}
The three same-context pairings assess reciprocal coordinate-frame reasoning, action--observation association, and behavior--visibility consistency, supporting relative-position navigation, action decision, view adjustment, and persistent target observation. Table~\ref{tab:dual_cognition_overall} exposes their different demands: Muse Spark\textsuperscript{T} leads overall Dual Semantic Acc (17.4\%), GLM-5.3-Flash\textsuperscript{T} is strongest across both image pairings among open-weight models, and Muse Spark\textsuperscript{T} also obtains the best proprietary behavior--visibility evidence score (16.2\%). Gemini~3.8 Flash\textsuperscript{T}, although outside its group's top five by overall semantics, leads the full model set in Dual Evidence Acc (13.2\%). These low and differently ranked scores show that strong marginal performance rarely extends to coordinated Agent-State and World-State success.
Muse Spark's action--observation scores (32.6\% semantic versus 3.4\% evidence success) and lower behavior--visibility semantic score (12.6\%) further reveal imbalances between semantic reasoning and grounding, and across paired capabilities. Reliable UAV agents require both capabilities across spatial and temporal tasks.
The full results are on the Analysis page.

\noindent \ding{47} \textbf{Human, thinking, and diagnostic validation.}
Human annotators reach 86.4\% image and 85.1\% video accuracy on stratified pools, supporting human solvability. \begin{figure*}[!t]
    \centering
    \includegraphics[width=\linewidth]{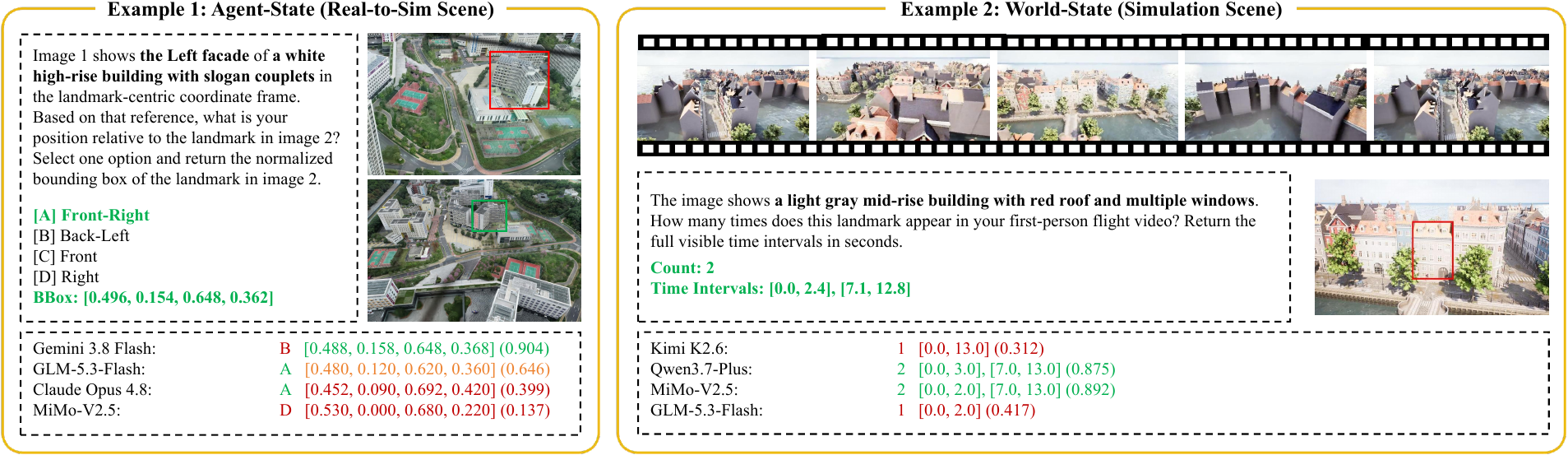}
    \caption{Representative dual-cognition cases. The Agent-State example separates relative-position semantics from BBox localization; the World-State example separates visibility counting from continuous interval localization.}
    \label{fig:qualitative}
\end{figure*}
 Controlled comparisons show that thinking is model- and modality-dependent: for Qwen3.7-Plus, it raises image/video accuracy from 41.7\%/31.4\% to 55.0\%/39.9\%; for MiMo-V2.5, it raises image accuracy from 40.3\% to 44.7\% but reduces image mIoU from 33.8\% to 24.5\% and video accuracy from 41.6\% to 34.9\%. Thus, additional deliberation can improve semantic decisions without consistently improving grounding or cross-media performance. Across 700 Instant and 600 Thinking failures, Instant errors concentrate on dual-coordinate reasoning (33.7\%), temporal tracking (23.6\%), and motion semantics (21.9\%). Full comparisons, protocols, and diagnostic distributions are on the Analysis page.

\noindent \ding{47} \textbf{Open-ended and qualitative validation.}
Removing answer choices reveals similar bottlenecks across 6 models and all 12 scenes. For Qwen3.5-9B, landmark-relative and self-relative position accuracy are 11.73\% and 29.88\%; atomic behavior reaches 62.55\% accuracy but only 26.10\% F1@0.5. Similar semantic and grounding difficulties persist in the open-ended setting. Fig.~\ref{fig:qualitative} further contrasts correct semantics with incomplete spatial or temporal grounding. Full results and matched task examples are on the Analysis page.

\noindent \ding{47} \textbf{Sim-to-real validation.}
We extend the construction pipeline to three CityGaussianV2~\cite{liu2025citygaussianv2} reconstructions built from real GauU-Scene~\cite{xiong2024gauuscenev2} aerial scenes across campus and downtown. After adapting the rendering bridge, we review 300 candidates, retain 30 landmarks, and generate 400 image questions (100 per task), on which we report 16 models. Table~\ref{tab:3dgs_dual_cognition} reproduces the broad trend observed in simulation: higher marginal semantic accuracy does not translate into strong dual-cognition coordination across paired Agent-State and World-State questions, while semantic and evidence rankings reveal distinct coordination bottlenecks. This persistent imbalance highlights the protocol's diagnostic value under real aerial appearance and geometry. Full construction and model-level results are on the Sim-to-Real page.

\begin{table}[t]
\centering
\caption{Two-stage optimization results on Qwen3.5-4B using UAV-DualCog-Train structured supervision.}
\label{tab:optimization_probe}
\scriptsize
\setlength{\tabcolsep}{3.5pt}
\renewcommand{\arraystretch}{0.90}
\begin{tabular}{lccc}
\toprule
\textbf{Metric} & \textbf{Base} & \textbf{Optimized} & \textbf{Gain} \\
\midrule
Image option Acc & 38.99\% & 59.77\% & +20.78 pp \\
Image BBox Acc@0.5 & 9.27\% & 61.04\% & +51.76 pp \\
Image mIoU & 20.12\% & 52.78\% & +32.66 pp \\
Video global Acc & 16.75\% & 26.67\% & +9.91 pp \\
Video F1@0.5 & 16.51\% & 33.40\% & +16.89 pp \\
Video mtIoU & 18.74\% & 34.58\% & +15.84 pp \\
\bottomrule
\end{tabular}
\end{table}

\subsection{Two-Stage Optimization Probe}
We further examine whether UAV-DualCog can serve as effective supervision beyond evaluation. UAV-DualCog-Train contains 11,232 image samples from six scenes disjoint from testing, covering four tasks, two difficulty levels, 234 landmarks, and the answer--BBox output contract. Using Qwen3.5-4B as the base model, multimodal supervised fine-tuning first stabilizes structured outputs; joint group relative policy optimization then optimizes
\begin{equation}
 R=R_{\mathrm{format}}+R_{\mathrm{answer}}+R_{\mathrm{bbox}},
\end{equation}
whose terms score schema validity, answer correctness, and normalized-box GIoU, respectively.
Table~\ref{tab:optimization_probe} shows gains in image option accuracy, BBox Acc@0.5, and mIoU. Image-only training also improves video global accuracy and temporal grounding overall, indicating cross-modal transfer of structured supervision. Dataset details, hyperparameters, reward definitions, full results, and optimization curves are on the Training page.

\section{Conclusion}
We introduced \textbf{UAV-DualCog}, a benchmark jointly evaluating Agent-State and World-State Reasoning across multi-view images and egocentric videos with explicit BBox and temporal localization. Paired Agent--Landmark contexts jointly assess coordinate-frame reasoning, action--observation association, and behavior--visibility localization. Broad model evaluation, paired analysis, open-ended validation, and the human baseline show that current MLLMs still struggle to answer both state questions and meet their localization requirements, especially over time. UAV-DualCog-Train further demonstrates the value of this structured supervision. Simulation provides diverse scenes and complete, controllable UAV poses, landmark geometry, visibility states, and 3D annotations for scalable construction and reproducible localization labels. Our three-scene 3DGS pilot extends the protocol to real aerial reconstructions and reproduces the broad capability trends observed in simulation. Future work will extend evaluation to physical UAVs across more diverse real-world scenes and develop models with stronger coordinated dual cognition and spatio-temporal grounding during embodied action.

\bibliographystyle{ieeetr}
%\IEEEtriggeratref{40}
\bibliography{content/references/references}

\end{document}